\documentclass{article}

\usepackage{amsmath}
\usepackage{amssymb}
\usepackage{amsthm}
\usepackage{amsfonts}       
\usepackage{authblk}
\usepackage{arxiv}
\usepackage{booktabs}       
\usepackage[utf8]{inputenc} 
\usepackage[nolist, nohyperlinks]{acronym}
\usepackage{enumitem}
\usepackage[T1]{fontenc}    
\usepackage{hyperref}       
\usepackage{url}            
\usepackage{nicefrac}       
\usepackage{microtype}      
\usepackage{graphicx}
\usepackage{subfig}
\usepackage{natbib}
\usepackage{doi}

\begin{acronym}
\acro{API}{Application Programming Interface}
\acro{DST}{Deep Sea Treasure}
\acro{ESR}{Expected Scalarized Return}
\acro{MaOO}{Many-Objective Optimization}
\acro{MDP}{Markov Decision Process}
\acro{ML}{Machine Learning}
\acro{MODRL}{Multi-Objective Deep Reinforcement Learning}
\acro{MOEA}{Multi-Objective Evolutionary Algorithm}
\acro{MOO}{Multi-Objective Optimization}
\acro{MORL}{Multi-Objective Reinforcement Learning}
\acro{RL}{Reinforcement Learning}
\acro{SER}{Scalarized Expected Return}
\acro{WFG}{Walking Fish Group}

\acrodefplural{MDP}[MDPs]{Markov Decision Processes}
\acrodefplural{MOEA}[MOEAs]{Multi-Objective Evolutionary Algorithms}
\end{acronym}

\newcommand{\newrlenv}[0]{TRI-OBJECTIVE DEEP SEA TREASURE}

\newcommand{\environmentURL}[0]{\href{https://github.com/imec-idlab/deep-sea-treasure}{https://github.com/imec-idlab/deep-sea-treasure}}
\newcommand{\environmentPackage}[0]{\href{https://pypi.org/project/deep-sea-treasure/}{https://pypi.org/project/deep-sea-treasure/}}
\newcommand{\environmentDOI}[0]{\href{https://doi.org/10.5281/zenodo.5227091}{https://doi.org/10.5281/zenodo.5227091}}
\newcommand{\acknowledgement}[0]{This research received funding from the Flemish Government (AI Research Program).\\This work was supported by the Research Foundation Flanders (FWO) under Grant Number 1SC8821N.\\The computational resources (Stevin Supercomputer Infrastructure) and services used in this work were provided by the VSC (Flemish Supercomputer Center), funded by Ghent University, FWO and the Flemish Government – department EWI.}

\title{A Review of the Deep Sea Treasure problem as a Multi-Objective Reinforcement Learning Benchmark}

\author[1]{Amber Cassimon}
\author[1]{Reinout Eyckerman}
\author[1]{Siegfried Mercelis}
\author[2]{Steven Latré}
\author[1]{Peter Hellinckx}

\affil[1]{University of Antwerp - imec \\
    IDLab - Faculty of Applied Engineering \\
    Sint-Pietersvliet 7, 2000 Antwerp, Belgium}
    
\affil[2]{University of Antwerp - imec \\
    IDLab - Department of Computer Science \\
    Sint-Pietersvliet 7, 2000 Antwerp, Belgium}

\date{}

\renewcommand{\headeright}{}
\renewcommand{\undertitle}{}
\renewcommand{\shorttitle}{A Review of the Deep Sea Treasure problem as a Multi-Objective Reinforcement Learning Benchmark}

\hypersetup{
pdftitle={A Review of the Deep Sea Treasure problem as a Multi-Objective Reinforcement Learning Benchmark},
pdfsubject={cs-LG},
pdfauthor={Amber Cassimon, Reinout Eyckerman, Siegfried Mercelis, Steven Latré, Peter Hellinckx},
pdfkeywords={Multi-Objective, Reinforcement Learning, Benchmark},
}

\begin{document}
\maketitle

\begin{abstract}
    In this paper, the authors investigate the \ac{DST} problem as proposed by Vamplew et al. Through a number of proofs, the authors show the original \ac{DST} problem to be quite basic, and not always representative of practical \ac{MOO} problems. In an attempt to bring theory closer to practice, the authors propose an alternative, improved version of the \ac{DST} problem, and prove that some of the properties that simplify the original \ac{DST} problem no longer hold. The authors also provide a reference implementation and perform a comparison between their implementation, and other existing open-source implementations of the problem. Finally, the authors also provide a complete Pareto-front for their new \ac{DST} problem.
\end{abstract}

\acresetall

\keywords{Multi-Objective \and Reinforcement Learning \and Benchmark}

\section{INTRODUCTION}
\label{sec:intro}
    \subsection{\acl{MORL}}
        \ac{MORL} is the subfield of \ac{RL} that attempts to find optimal policies for problems with at least two objectives. Within this field, various techniques have been proposed, such as \acp{MOEA}\cite{Deb2002}, Deep Q-Learning with Lexicographical Thresholding~\cite{Hayes2020} and Outer-Loop Approaches like Deep OLS~\cite{Mossalam2016}. An essential property of \ac{MORL} or even \ac{MOO} in general is that each of the objectives must be in conflict with the others. When two objectives can be optimized concurrently, the problem becomes equivalent to the problem of optimizing the sum of both objectives, effectively reducing the number of objectives by one. The problem discussed in this paper consists of 2 to 3 conflicting objectives, and can thus be classified as a \ac{MORL} problem.

    \subsection{\acl{MaOO}}
        While there is no exact definition of when a \acl{MOO} problem becomes a \ac{MaOO} problem, the line is typically drawn around 4 objectives, with anything with more than 1, but less than 4 objectives considered a \acl{MOO} problem. \cite{Saxena2009} define a \ac{MaOO} problem to have `significantly more than five' objectives, while \cite{Cheng2016}, \cite{Mane2017}, \cite{Wang2015} and \cite{Yang2013} say \ac{MaOO} problems are any problem with more than three objectives. Despite these differences, it is generally accepted that \ac{MaOO} problems are significantly more challenging and less intuitive than \ac{MOO} problems. While the authors thought it important to point out the difference between \ac{MOO} and \ac{MaOO}, this paper will restrict itself to \ac{MOO} problems.

    \subsection{\aclp{MDP}}
        \acp{MDP} are the most commonly used way of describing problem statements in the field of \acl{RL}. An \ac{MDP} is usually described as a 4-tuple \cite{Sutton1998}:

        $$MDP = \left(\mathcal{S}, \mathcal{A}, p\left(s' | s, a\right), \mathcal{R}\left(s, a\right)\right)$$

        The elements of the \ac{MDP} are the state-space ($S$), the action-space ($A$), the dynamics ($ p\left(s' | s, a\right)$)  and the reward function ($\mathcal{R}\left(s, a\right)$).
        The state-space describes the environment's state. Some \ac{RL} problems are fully-observable, meaning that the agent gets to observe the complete state of the environment, while others are partially observable, meaning that the agent only gets to observe part of the state of the environment.
        The action-space determines the actions which an agent can take while operating in an environment. It is possible for this action space to be dependent on the current state of the environment, in which case it is usually denoted as $\mathcal{A}\left(s\right)$. In this paper however, this is not the case, so we will simply denote the action space as $\mathcal{A}$.
        The dynamics of an environment ($p\left(s' | s, a\right)$) determine how the environment evolves and reacts to an agent's actions. Essentially, it is a function that gives the probability distribution of new states, conditional on the current state, and the action which the agent took.
        Finally, there is the reward function $\mathcal{R}\left(s, a\right)$. In most \ac{RL} literature, the reward is defined as
        $$\mathcal{R}: \left(S, A\right) \in \mathcal{S} \times \mathcal{A} \mapsto \mathbb{R}$$
        Unfortunately, when dealing with \ac{MORL} this formulation requires some changes, since our environments return a reward vector consisting of $N$ objectives, resulting in a definition of
        $\mathcal{R}$ as $$\mathcal{R}: \left(S, A\right) \in \mathcal{S} \times \mathcal{A} \mapsto \mathbb{R}^{N}$$ with $N \in \left\{2, 3, \ldots\right\}$.
        
        In section~\ref{sec:sota}, the authors discuss existing \ac{MOO} and \ac{MORL} benchmarks, and open-source \ac{DST} implementations. This is followed by a detailed description of the original \ac{DST} problem proposed by Vamplew et al.\cite{Vamplew2011} in section~\ref{sec:dst}. After discussing the existing problem, the authors showcase their improved version of the problem in section~\ref{sec:newdst}, and followed by a discussion of their implementation~\ref{sec:impl}. Finally, the authors provide some potential directions for future research in section~\ref{sec:discussion-future_work}.

\section{STATE OF THE ART}
\label{sec:sota}
    \subsection{Optimality Criteria}
    \label{ssec:sota:criteria}
        When considering scalarizing \ac{MORL} approaches, there are two different ways of performing this scalarization, and each of these techniques corresponds to a different notion of a good solution. In a recent survey paper on utility-based \ac{MORL}, \cite{Radulescu2019} described these two optimality criteria: \ac{ESR} and \ac{SER}. In their paper, they formulate \ac{ESR} as shown in equation~\ref{eq:sota:esr}, where $u()$ is the scalarization function:
        \begin{equation}
            V_{u}^{\pi} = \mathbb{E}\left[u\left(\sum_{t=0}^{\infty}{\gamma^{t}r_{t}}\right)| \pi, \mu_{0}\right]
            \label{eq:sota:esr}
        \end{equation}
        
        The mathematical formulation of \ac{SER}, on the other hand, is shown in equation~\ref{eq:sota:ser}
         \begin{equation}
            V_{u}^{\pi} = u\left(\mathbb{E}\left[\sum_{t = 0}^{\infty}{\gamma^{t}r_{t} | \pi, \mu_{0}}\right]\right)
            \label{eq:sota:ser}
        \end{equation}
       
        Under \ac{ESR}, a user derives utility from a single roll-out of its policy, while under \ac{SER} utility is derived from the expected outcome (mean over multiple roll-outs). In the context of \ac{MOO}, the notions of \ac{ESR} and \ac{SER} are very valuable, since they describe two fundamentally different types of solutions. A practical example of \ac{ESR} is the LEMONADE algorithm, from Elsken et al.~\cite{Elsken2018}. Using LEMONADE, Elsken et al. find the optimal neural network architecture for image classification on CIFAR-10 and ImageNet64x64. In this paper, the architecture search problem is formulated as an \ac{ESR} problem, since the author's policy (an evolutionary algorithm) is only executed once to obtain results. An example of \ac{SER} is demonstrated by Khamis et al. in their paper on \ac{MORL} for traffic signal control~\cite{Khamis2014}. The policy found by Khamis et al. is executed many times, and is only optimal if it can consistently route traffic in an optimal way (\ac{SER}), rather than generating a single optimal traffic light configuration (\ac{ESR}).
    
    \subsection{Existing Benchmark Suites}
    \label{ssec:sota:benchmarks}
        In \ac{MOO} Literature, there exist numerous benchmarks, such as the \ac{WFG} Toolkit~\cite{Huband2005}, ZTD Benchmark~\cite{Zitzler2000}, DTLZ Benchmark~\cite{Deb2001}. Vamplew et al.~\cite{Vamplew2011} also included a link to a list of benchmarks in their original publication. Unfortunately, at the time of writing, the given link\footnote{\href{http://hdl.handle.net/102.100.100/4461}{http://hdl.handle.net/102.100.100/4461}} no longer works. When analyzing benchmarks like ZTD, DTLZ and WFG, as well as the problems proposed in~\cite{Vamplew2011}, we notice that almost all of them are tailored towards \ac{ESR}-style optimization. While this provides a nice playground for \acp{MOEA}, it doesn't suit many \ac{RL} problems well, since they are usually formulated under the \ac{SER} criterion, as opposed to \acp{MOEA}, which are usually built around the concept of \ac{ESR} (An example of \ac{RL} being used with an \ac{ESR} criterion is the architecture search performed by Pham et al.~\cite{Pham2018}).
    
    \subsection{\acl{DST}}
    \label{ssec:sota:dst}
        One commonly occurring \ac{MORL} benchmark problem is the \acf{DST} problem. In this problem, a submarine is tasked with collecting treasures. For each timestep the submarine spends collecting treasures, it is penalized with a time-score of -1. Every step that the submarine spends not finding a treasure, it is given a treasure score of 0, and when the submarine finds a treasure, it is given a treasure score equal to the value of the treasure. This problem was originally proposed by Vamplew et al.~\cite{Vamplew2011} as part of an effort to provide a benchmark suite for \ac{MORL} algorithms. Since its introduction, the \ac{DST} problem has been used numerous times as a benchmark problem in \ac{MORL} literature \cite{VanMoffaert2014a}, \cite{VanMoffaert2013}, \cite{Mossalam2016}
        
    \subsection{Open Source Implementations}
    \label{ssec:sota:impls}
        When benchmarking \ac{RL} algorithms, a commonly used \ac{API} is that of OpenAI's gym~\cite{Brockman2016}. Gym proposes a simple software \ac{API} that encapsulates the underlying \ac{MDP}. Through this \ac{API}, gym allows for algorithms to be easily tested on multiple different problems to show an algorithms capability to generalize to different problems.\\
        A number of open-source implementations of the \ac{DST} problem exist. Van Moffaert and Nowé provide an implementation of the \ac{DST} problem as part of their paper on Pareto-Q Learning~\cite{VanMoffaert2014a}. Their implementation\footnote{\href{https://gitlab.ai.vub.ac.be/mreymond/deep-sea-treasure}{https://gitlab.ai.vub.ac.be/mreymond/deep-sea-treasure}} is written in Python, and compliant with the gym interface. This implementation does not allow for arbitrary Pareto-fronts to be used, although they do provide an implementation of the Bountyful Sea Treasure variant~\cite{VanMoffaert2014b}.
        Another implementation is that of Nguyen et al.~\cite{Nguyen2018}, as part of their \ac{MODRL} framework. Their framework contains both problem statements and solution strategies. The framework\footnote{\href{https://personal-sites.deakin.edu.au/~thanhthi/drl.htm}{https://personal-sites.deakin.edu.au/~thanhthi/drl.htm}} is written in Python. The environment isn't fully compatible with the OpenAI gym interface (The implementation is missing certain attributes such as \texttt{action\_space} and \texttt{observation\_space}, and likely doesn't implement all functionality present in a regular gym environment, since it doesn't inherit from \texttt{gym.Env} class or any of its subclasses). While the \ac{API} doesn't match that of gym, it does present an interface that is similar enough to allow research to rapidly adapt it to a gym-compatible environment. This implementation is much more customizable, and while it doesn't allow for the specification of arbitrary Pareto-fronts, it does have the built-in capability to present a convex, concave, linear and mixed (convex and concave) Pareto-front.

\section{BI-OBJECTIVE DEEP SEA TREASURE}
\label{sec:dst}
    First, we will analyze the original \ac{DST} Problem, as proposed by \cite{Vamplew2011}. We will start this discussion by defining the exact \ac{MDP} that the \ac{DST} Problem proposes. In the spirit of keeping this paper compact, theoretical proofs a basic set of properties of the \ac{DST} problem and its solutions are omitted from this paper, and presented in the supplementary material.
    
    The \ac{MDP} that defines the \ac{DST} Problem is a finite-horizon \ac{MDP} (Vamplew et al. originally limited the environment to 1000 time steps~\cite{Vamplew2011}) with the state and action spaces defined in equation~\ref{eq:dst:mdp}. The dynamics and reward function will be elaborated further in section~\ref{ssec:dst:dynamics} and section~\ref{ssec:dst:rewards} (We follow Sutton and Barto's convention of using $\doteq$ for definitions~\cite{Sutton1998}).
    
    \begin{equation}
        MDP_{2DST} \doteq \left(\left\{0 \ldots 9\right\} \times \left\{0 \ldots 10\right\}, \left\{0, 1, 2, 3\right\}, p\left(s' | s, a\right), \mathcal{R}\left(s, a\right)\right)
        \label{eq:dst:mdp}
    \end{equation}
    
    Through this section, and section~\ref{sec:newdst}, we will use the letter $t$ to denote time.
    
    \subsection{State Space}
    \label{ssec:dst:state-space}
        Looking at the state-space from equation~\ref{eq:dst:mdp}, we can see the agent observes 2 positive integers, representing the agent's current position as an $(x, y)$ pair. We note that the agent starts at the point with coordinate $(0, 0)$ in the top-left corner of the environment, and as the agent moves left-to-right, and top-to-bottom, its respective x and y coordinates increase. The original DST problem consisted of a $10 \times 11$ grid, resulting in a maximum state space size of 110 (The actual size of the state space is 60, since not all squares of the grid are accessible to the agent).
    
    \subsection{Action Space}
    \label{ssec:dst:action-space}
        The agent has 4 potential actions, corresponding to each of the cardinal directions in the world, enumerated in clockwise fashion (UP, RIGHT, DOWN, and LEFT). Given the state space of size 60, and an action space of size 4, we can determine that a solution using tabular Q-learning would only require a table with around 240 entries. This is a first indicator of the simplicity inherent in the original \ac{DST} problem, and presents a strong argument in favour of more complicated problems when using Deep Learning-based techniques.
     
    \subsection{Dynamics}
    \label{ssec:dst:dynamics}
        The environment dynamics for the \ac{DST} problem are completely deterministic:
        \begin{equation}
            s \doteq \left(x_{t}, y_{t}\right)
        \end{equation}
        
        \begin{align}
            \Delta x\left(a\right) \doteq 
            \begin{cases}
                1   & \mbox{if } a = 1\\
                -1  & \mbox{if } a = 3\\
                0   & \mbox{otherwise}
            \end{cases}
        \end{align}
    
        \begin{equation}
            x_{t + 1}\left(s, a\right) \doteq
            \begin{cases}
                x_{t} + \Delta x\left(a\right) & \mbox{if } \mbox{ collides}\left(x_{t} + \Delta x\left(a\right)\right) = 0\\
                x_{t} & \mbox{otherwise}
            \end{cases}
            \label{eq:dst:collides}
        \end{equation}
        
        \begin{align}
            p\left(s' | s, a\right) =
            \begin{cases}
                1 & \mbox{if } s' = \left(x_{t + 1}\left(s, a\right), y_{t + 1}\left(s, a\right)\right)\\
                0 & \mbox{otherwise}
            \end{cases}
        \end{align}
    
    The $collides\left(\right)$ function in equation~\ref{eq:dst:collides} returns an integer in $\left\{0, 1\right\}$ indicating if the new position would cause a collision or not. We only describe the procedure for obtaining the next x-value from the current x value and the action, but the procedure for obtaining the next y-value is analogous.

    \subsection{Rewards}
    \label{ssec:dst:rewards}
        The reward function in the original \ac{DST} problem returns a 2-dimensional reward, consisting of a time component, and a treasure component. We denote the time component as $r_{b}\left(s, a\right)$ and the treasure component as $r_{p}\left(s, a\right)$ (P for plunder), resulting in the following reward function:
        $$\mathcal{R}\left(s, a\right) = \left(r_{b}\left(s, a\right), r_{p}\left(s, a\right)\right)$$
        with
        \begin{equation}
            r_{b}\left(s, a\right) \doteq -1
        \end{equation}
        and
        \begin{equation}
            r_{p}\left(s, a\right) \doteq 
            \begin{cases}
                p \in \mathbb{R}^{+}_{0} & \mbox{in terminal state}\\
                0 & \mbox{if state is not terminal}
            \end{cases}
        \end{equation}
        
        The exact value of the treasures in the \ac{DST} environment can vary between publications. Vamplew et al. proposed an original set of treasures in their paper~\cite{Vamplew2011}, Mossalam et al. proposed an alternative set of treasures to make the problem convex in their publication~\cite{Mossalam2016}. Another set of treasure values can be found in \cite{VanMoffaert2014b}, where the problem was renamed to Bountiful Sea Treasure. For the remainder of this publication, we will not consider any particular set of treasure values, rather, we will simply assume that all treasure values are $>0$.
        
        In the bi-objective \ac{DST} environment, it can be proven that actions causing a collision always result in a solution which is Pareto-dominated by the same solution where the colliding action is removed. It can also be shown that any solution that results in a time-out, rather than the agent finding a treasure will be Pareto-dominated by a solution that finds a treasure. For proof of these properties, we refer to the supplementary material included with this paper. Using these properties we can exclude certain solutions from the search space, making it substantially smaller.

\section{\newrlenv}
\label{sec:newdst}
    In this section, we will introduce an improved alternative to the original \ac{DST} problem. This version of the \ac{DST} problem involves the optimization of three conflicting objectives: Time, Treasure and Fuel, and has a larger observation and action space than the original \ac{DST} problem. Through the changes in observation space, the new \ac{DST} problem more closely resembles an \ac{SER} problem than the original \ac{DST} problem, bringing it more in-line with traditional \ac{RL} problems. Similar to the \ac{DST} problem, we define this problem to be a finite-horizon \ac{MDP}, limited to 1000 time steps.

    \begin{equation}
        MDP_{3DST} \doteq (\mathbb{Z}^{2 \times 11}, \left\{-3, -2, \ldots, 2, 3\right\}^{2}, p\left(s' | s, a\right), \mathcal{R}\left(s, a\right))
    \end{equation}
    
    \subsection{State Space}
    \label{ssec:newdst:state-space}
        The state space of this new \ac{DST} problem is significantly larger than the old one. The new state space consists of eleven two-element column vectors. The first column vector represents the agent's current velocity, expressed as a separate x- and y-component. The following ten column vectors each represent the agent's relative coordinates to each of the treasures. This observation differs from the original \ac{DST} problem in that it directly tells the agent where each of the potential solutions are. By changing the formulation of the observation space, the agent's task changes from memorizing the treasure locations and a path to them, to learning how to traverse the ocean towards whatever treasure the agent finds more interesting. Through this reformulation of the problem, the optimality criterion for the \ac{DST} problem can also shift to \ac{SER}, rather than \ac{ESR} if, for example, treasure locations were randomized. When analyzing this environment's state space, we note that the final ten columns of the observation all capture the same state (the agent's position). From section~\ref{ssec:dst:state-space}, we know that there are 60 positions the agent can visit. In the tri-objective environment, the agent's velocity is normally capped at 5 in any direction, meaning that there are a total of 121 different velocity vectors the agent can achieve ($\left(5 + 1 + 5\right)^2$).  While not every velocity vector is achievable in every position, this still allows us to place an upper bound on the size of our state-space of 7260, which is already significantly larger than the original \ac{DST} problem, which had a state-space of 60 elements.
    
    \subsection{Action Space}
    \label{ssec:newdst:action-space}
        The action-space consists of a two-element vector. The elements of this vector represent the respective x- and y-components of the acceleration the agent would like to make. In the new \ac{DST} environment, the agent no longer takes discrete steps in one of the cardinal directions, but rather, the agent can accelerate in one or both dimensions, allowing for more efficient ways of reaching each treasure at the cost of a higher fuel consumption. The total size of our action space is 49 ($7 \times 7$). When combining this with our state-space, we get a combined state-action space of $355740$ elements. While this is an upper bound, and the actual number will likely be lower, it is still a significant increase over the original \ac{DST} problem, which had a state-action space of 240 elements.
    
    \subsection{Dynamics}
    \label{ssec:newdst:dynamics}
        At a high-level, the dynamics of the new \ac{MDP} are similar to that of the old \ac{MDP}: If the agent attempts to make a move that would result in a collision, the state if left unchanged, otherwise, the agent's desired action is executed. Similar to section~\ref{ssec:dst:dynamics}, we will only show the dynamics function for the x-component for brevity, noting that a set of identical operations are executed for the y-component.
        We start off by defining our action $a$:
        
        \begin{equation}
            a \doteq \left(a_{x}, a_{y}\right)
        \end{equation}
        
        Next, we determine a preliminary next velocity and position, which will be used for collision checking.
        
        \begin{align}
            v'_{t + 1, x} &= v_{t, x} + a_{x}\\
            x'_{t + 1} &= x_{t} + v'_{t + 1, x}
        \end{align}
        
        Knowing the position of the submarine if the action was executed, we can check for collisions, and update our actual velocity and position accordingly.
        
        \begin{equation}
            v_{t + 1, x} = 
            \begin{cases}
                v_{t, x} + a_{x} &\mbox{ if } collides\left(x'_{t + 1}\right) = 0\\
                0 &\mbox{ otherwise }
            \end{cases}
        \end{equation}
        
        \begin{equation}
            x_{t + 1} = x_{t} + v_{t + 1, x}
        \end{equation}
        
        We note that in this case, the $collides()$ function should not only check for collisions in the cardinal directions, but also in diagonal directions. Another important note is that the agent's velocity is reduced to zero if an action would cause a collision, which is a behaviour an agent could attempt to exploit to arrest movement without consuming fuel.
    
    \subsection{Rewards}
    \label{ssec:newdst:rewards}
        In terms of rewards, the new \ac{DST} problem inherits the two objectives from the old \ac{DST} problem (time and treasure), and adds a third objective, fuel. The fuel objective was designed in such a way that it conflicts with the two already existing objectives. When the agent attempts to find far-away treasures, it will need to spend more fuel to cover the distance, and when the agent wants to cover a given distance quicker, it can spend more fuel accelerating and decelerating.
        
        Formally, we define the fuel objective $r_{f}$ (F for fuel) as the sum of the squares of the acceleration in both dimensions, if the action does not cause a collision, or 0 if the action would cause a collision:
        
        \begin{equation}
            r_{f} =
            \begin{cases}
                -\left(a_{x}^{2} + a_{y}^{2}\right) &\mbox{ if } collides\left(x'_{t + 1}\right) = 0\\
                0                                               &\mbox{ otherwise }
            \end{cases}
        \end{equation}
    
        We note that this function creates the exploit mentioned earlier in section~\ref{ssec:newdst:dynamics}. The formulation of this reward function also does not necessarily make collisions sub-optimal, since coasting (Moving without accelerating using built-up inertia) uses the same amount of fuel as colliding.
        
        This leads us to $\mathcal{R}\left(s, a\right)$ as:
        
        \begin{equation}
            \mathcal{R}\left(s, a\right) = \left(r_{b}\left(s, a\right), r_{p}\left(s, a\right), r_{f}\left(s, a\right)\right)
        \end{equation}

    For the tri-objective \ac{DST} problem, we show that there exists at least 1 solution ending in a time-out that is part of the Pareto-front. The proof for this property can be found in the supplementary material. Besides this, the authors also attempt to prove whether or not a collision impacts the optimality of a solution. Unfortunately, the new acceleration-based action space makes this proof significantly more complicated than for the bi-objective environment. Following a line of reasoning similar to that used for this proof in the bi-objective case, it quickly becomes clear that this can not be proven in the same way. In the tri-objective case, a collision necessarily must be considered as consisting of multiple timesteps, since it is possible for an agent to accumulate more velocity than it can arrest in a single step. Proving or disproving this property would thus require a fundamentally different line of reasoning, which the authors feel that this speaks to the added complexity inherent in the tri-objective environment.

\section{IMPLEMENTATION}
\label{sec:impl}
    The authors also provide a reference implementation of both the new and old \ac{DST} problem, written in Python\footnote{\environmentURL}. The authors also published their code as a PyPI package\footnote{\environmentPackage}. The implementation provides numerous options. The proofs in the supplementary materials only consider the default options, since changes to these options would change the properties of the environment, but we believe they would still be valuable for researchers regardless. When building the implementation, care was taken to ensure the implementation conforms to the gym \ac{API}, to ease adoption among \ac{RL} practitioners.\\
    
    \subsection{Configurability}
    \label{ssec:impl:config}
\begin{figure}[h]
            \centering
            \subfloat[Custom Treasures and No Debug Information]{{\includegraphics[width = 0.21 \textwidth]{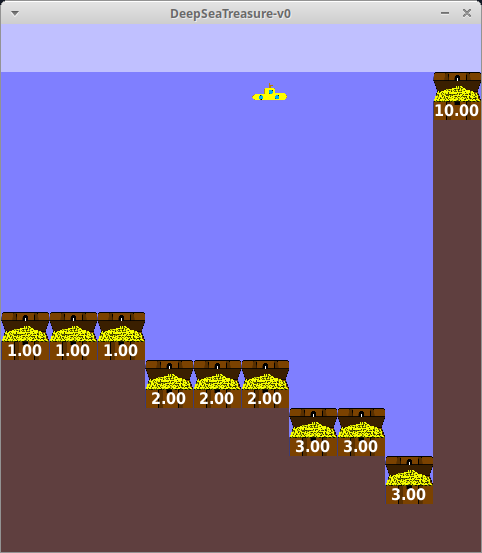} }}%
            \qquad%
            \subfloat[Default Treasures and Debug Information]{{\includegraphics[width = 0.21 \textwidth]{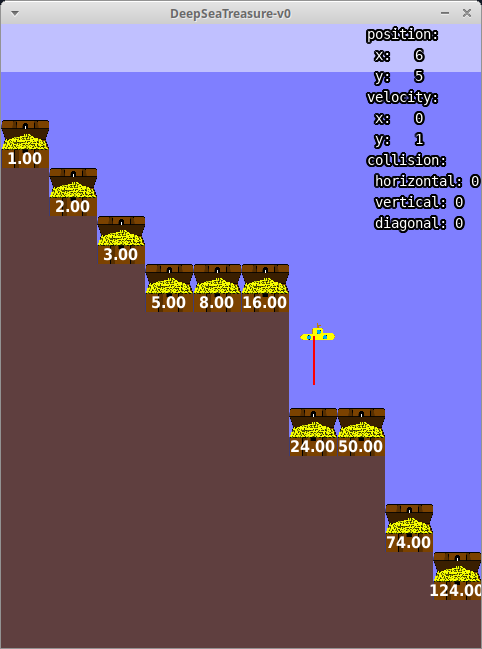} }}%
            \caption{Submarine moving through \ac{DST} environment}
            \label{fig:impl:debug-info}
        \end{figure}

        The original \ac{DST} problem proposed by Vamplew et al.~\cite{Vamplew2011} can easily be recreated using a separate wrapper in combination with the regular environment. The fuel objective is not an inherent part of the environment, but rather can be added by utilizing a different wrapper. Through the use of gym's wrapper system, the authors hope to provide a sufficiently composable environment that can be used for future research.\\
        The implementation allows its users to re-define the action space by specifying the positive acceleration levels in 1 dimension that the agent can use, along with this, the fuel costs for each acceleration level can also be set, allowing the user to completely re-define their fuel objective based on their preferences.\\
        The default treasure location and value are set to match those described by Vamplew et al.~\cite{Vamplew2011}, but can be reconfigured. This serves two functions, first, it allows users to redefine their Pareto-fronts for time and treasure, by moving treasures closer or further away and changing their exact treasure values. Secondly, it also allows users to build \ac{RL} agents that do not converge to a single solution, but rather that learn to search for and find solutions, bringing the optimization criterion closer to an \ac{SER}-like formulation, rather than an \ac{ESR}-like formulation.
        An example of a custom set of treasures is shown in Fig.~\ref{fig:impl:debug-info}a.\\
        In some cases, it may be interesting to intrinsically disregard collisions. While they can easily be proven to be sub-optimal in the bi-objective case, this is not true for the tri-objective case. The authors' implementation allows users to make collisions guaranteed to be sub-optimal, however. By setting the \texttt{implicit\_collision\_constraint} option, the environment will return a vector of immediate rewards that are worse than any other set of possible reward values.\\
        The implementation also has some additional functionality to help with debugging, such as the capability to display debug information, shown in Fig. ~\ref{fig:impl:debug-info}b. The environment provides many more options. For a detailed overview, we refer to the implementations documentation.
        
    \subsection{Pareto-Front}
    \label{ssec:impl:paretofront}
\begin{figure}
            \centering
            \includegraphics[width = 0.52 \textwidth]{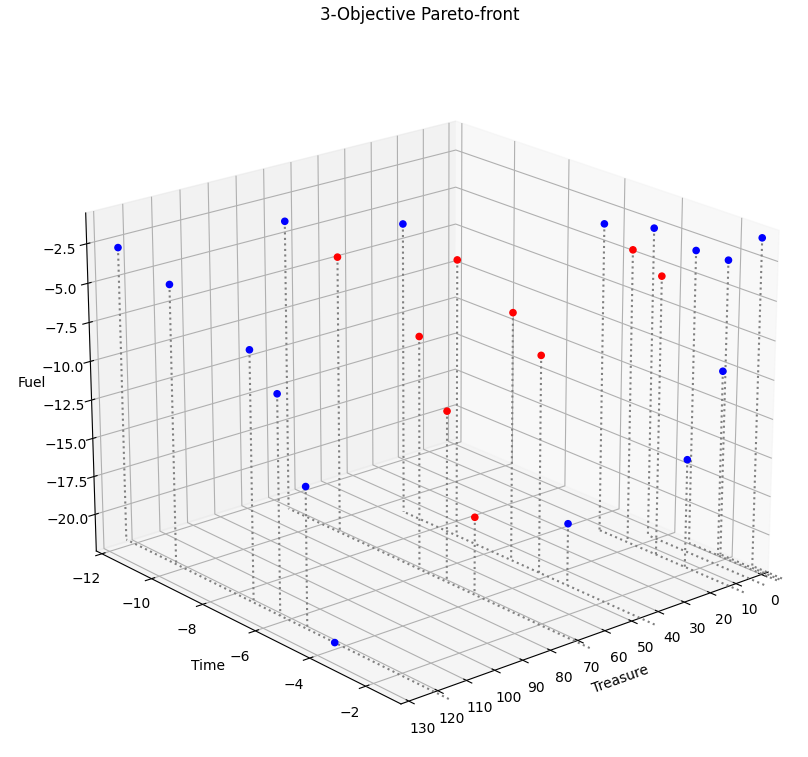}
            \caption{3-Objective Pareto-front. Points in blue lie on the convex hull, while points in red are contained inside the convex hull.}
            \label{fig:pf}
        \end{figure}

        Finally, we provide the Pareto-front for the tri-objective environment. While we do not discuss it here, the Pareto-front data for the bi-objective environment is also available in our repository. While we were unable to prove certain properties of the tri-objective environment, we were able to empirically determine the Pareto-front of the tri-objective environment through exhaustive graph search. Solving this environment took 33.22 hours using 28 Intel E5-2680v4 Broadwell CPUs. The numerical data for the Pareto-front can be found in the same repository that contains the code for the environment. Figure~\ref{fig:pf} shows the obtained Pareto-front for the tri-objective environment. It is important to note that the solution resulting in a time-out reported in section~\ref{sec:newdst} was omitted for clarity, it is however present in the numerical dataset in the repository. Each point on the Pareto-front has been projected onto the time-treasure plane for clarity, and horizontal lines have been drawn for all solutions that lead to the same treasure.
        
        \begin{figure}[h]
            \centering
            \includegraphics[width = 0.52 \textwidth]{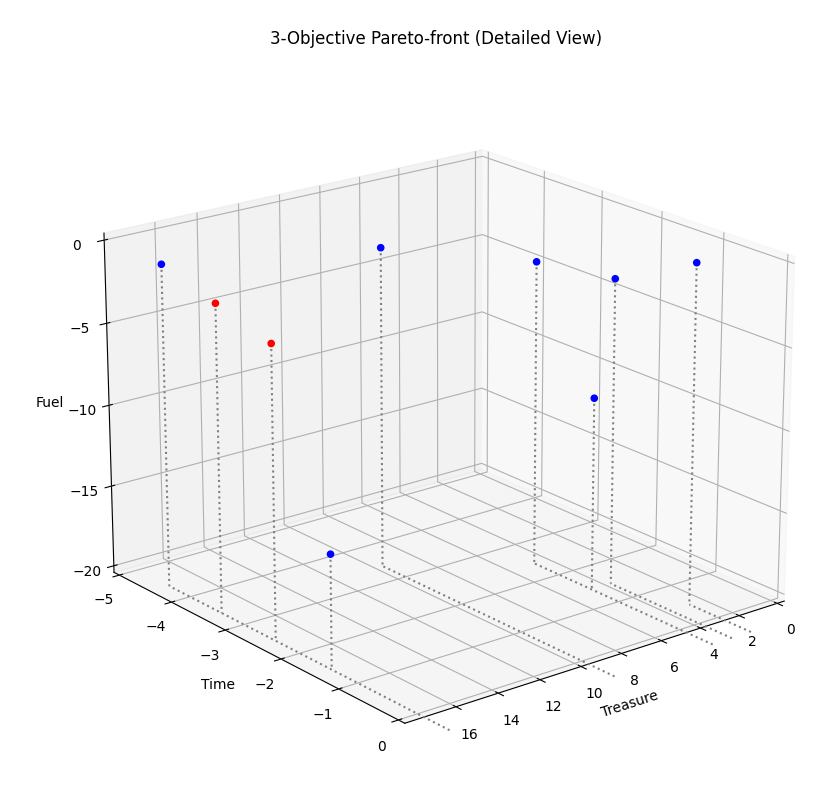}
            \caption{Detailed view of the low-treasure region of the 3-Objective Pareto-front. Points in blue lie on the convex hull, while points in red are contained inside the convex hull. A convex hull is the smallest, convex set of points that contains a given set of points.}
            \label{fig:pfdetail}
        \end{figure}
        
        Looking at Figure~\ref{fig:pf}, a first notable property of this Pareto front is that the number of points on the Pareto-front (25 + 1) no longer corresponds to the number of treasures (10). In our analysis, we will ignore the 26th point generated by the submarine performing the ``idle" action 1000 times. While this point is theoretically optimal, we argue that it does not produce a useful solution, and hence will disregard it in further analysis. We can see that for each treasure, there exist different trade-offs between fuel and time. We also note, that not all treasures are part of the Pareto-front, the treasures that make up the Pareto-front are those with values $\left\{1, 2, 3, 8, 16, 50, 74, 124\right\}$, while the treasures with values $\left\{5, 24\right\}$ are not part of the Pareto-front, as can be seen in the detailed view of the Pareto-front in~\ref{fig:pfdetail}.\\
        By looking at Figure~\ref{fig:pf} it can also be easily ascertained that the Pareto-front for this environment is not convex. In this Pareto-front, 9 out of 25 solutions lie inside the convex hull, rather than on it. In the original \ac{DST} problem, 3 out of 10 points formed local concavities. Similar to the original \ac{DST} problem, this presents an interesting challenge for \ac{MORL} algorithms, testing their ability to handle non-convex Pareto-fronts.
        
    \subsection{Guidelines}
    \label{ssec:impl:guidelines}
        While this implementation is highly configurable, the authors recommend anyone using their implementation to always report performance on both default configurations (\texttt{VamplewWrapper + DeepSeaTreasureV0} and \texttt{FuelWrapper + DeepSeaTreasureV0}) While it may be interesting to alter the default configuration in the context of a specific paper, using a non-standard configuration makes comparing the used algorithm to that of other researchers very difficult, or even impossible. Some options (like \texttt{render\_grid} or \texttt{render\_treasure\_values}) do not impact the found solutions, but other options can have a profound impact on the solutions that are found, and how quickly/easily they can be found (options such as \texttt{implicit\_collision\_constraint} and \texttt{treasure\_values}). We encourage users to make maximal use of the configurability offered by the environment, we also advocate for always reporting performance on custom configurations in addition to the default ones, to allow for fair comparisons between algorithms.\\
        When utilizing custom configurations, we also believe it to be critical to report the complete configuration of the environment. The authors have made sure to provide a simple way to report an environments' configuration through the use of the \texttt{config()} method. This method returns the complete configuration of the environment, and can be used to create a new environment with an identical configuration. The authors encourage anyone using this environment to always report the complete configuration of their environment by including this data in a \texttt{.json} file in the supplementary materials or appendix of the publication. The use of machine-readable format like JSON over a human-readable format like a table in a PDF file makes it easy for researchers aiming to reproduce each others work to easily copy and paste a configuration from a publication into their own code.
        The implementation and Pareto-front dataset provided by the authors is independently citable through Zenodo under DOI \environmentDOI.

\section{DISCUSSION \& FUTURE WORK}
\label{sec:discussion-future_work}
    One major omission in this paper is the concept of constraints. Constraints in the context of \ac{MOO} are usually formulated as a set of (in)equalities that check certain properties of a solution. In the case of the \ac{DST} problem, a constraint could be ``The agent is not allowed to visit square (5, 3)". While this may seem arbitrary in the context of the \ac{DST} problem, constraints are often found in real-life engineering applications, such as in the placement of distributed tasks in fog environments~\cite{Eyckerman2020} or the automatic design of neural networks targeting embedded devices~\cite{Cassimon2020}. Constraints also pose an interesting challenge to the \ac{API} proposed by gym~\cite{Brockman2016}, since the current \ac{API} provides no natural way of indicating that certain solutions are ``unacceptable" or ``invalid". While it is usually possible to work around this, by ending episodes early, and modifying reward functions to make constraint violations guaranteed to be sub-optimal, this is not always easy to achieve, and prompts the question of whether or not gym provides the most well-suited \ac{API} for solving constrained optimization problems using \ac{RL} agents.\\
    While much research has been done on solving \ac{MORL} problems under \ac{ESR} optimality criteria, with both evolutionary and \ac{RL} techniques, literature on algorithms targeting \ac{SER} optimality criteria is relatively sparse, even though this is the dominant problem formulation in \ac{RL}. With this in mind, the authors believe this to be a valuable research direction, since many \ac{RL} problems whose formulation is simplified to single-objective problems currently, could be revisited, while examining their complete solution space.\\
    The authors would also like to point out that, even though the complexity of the original \ac{DST} problem was significantly increased in the new, tri-objective \ac{DST} problem, the \ac{DST} problem should still be considered a toy problem.  Problems like Multi-Objective traffic signal control, like the one tackled by Khamis et al. ~\cite{Khamis2014} could provide valuable use-cases to show the suitability of \ac{MORL} algorithms in more practical use-cases.\\
    In this paper, the authors proposed a new environment, and tackled it from both a theoretical and practical perspective, providing a working, powerful and easy-to-use implementation. In order to reduce the scope of this paper, the authors decided against the execution of various \ac{MOO} and \ac{MORL} algorithms on the environment. While the comparison between algorithms and environments would undoubtedly provide valuable data and insights, the authors feel like this would be better suited for a separate publication.

\section*{ACKNOWLEDGEMENTS}
\acknowledgement

\bibliographystyle{apalike}
\bibliography{references} 

\end{document}